# AN AUTOMATED SIZE RECOGNITION TECHNIQUE FOR ACETABULAR IMPLANT IN TOTAL HIP REPLACEMENT

A. Shapi'i[1], R. Sulaiman[2], M.K. Hasan[3] and A.Y.M. Kassim[4]

[1]Industrial Computing Research Group, Universiti Kebangsaan Malaysia, Malaysia
`azrul@ftsm.ukm.my`
[2]School of Information Technology, Universiti Kebangsaan Malaysia, Malaysia
`rs@ftsm.ukm.my`
[3]Information Technology Center, Universiti Kebangsaan Malaysia, Malaysia
`khatim@ftsm.ukm.my`
[4]Department of Orthopaedic and Traumalogy, Medical Center of Universiti Kebangsaan Malaysia
`dryazidk@gmail.com`

*ABSTRACT*

*Preoperative templating in Total Hip Replacement (THR) is a method to estimate the optimal size and position of the implant. Today, observational (manual) size recognition techniques are still used to find a suitable implant for the patient. Therefore, a digital and automated technique should be developed so that the implant size recognition process can be effectively implemented. For this purpose, we have introduced the new technique for acetabular implant size recognition in THR preoperative planning based on the diameter of acetabulum size. This technique enables the surgeon to recognise a digital acetabular implant size automatically. Ten randomly selected X-rays of unidentified patients were used to test the accuracy and utility of an automated implant size recognition technique. Based on the testing result, the new technique yielded very close results to those obtained by the observational method in nine studies (90%).*

## 1. INTRODUCTION

The field of orthopaedics at the time became more important as the number of patients suffering from osteoporosis increased every year [1]. According to experts from PPUKM, the conventional detection method was used to search for a suitable implant for the patient. The method involves using the implant templates supplied by the supplier, and then the image of the implant is measured by doing image mapping on a patient's scanned X-ray. This step was used repeatedly until the appropriate implant of the patient encountered. However, the survey found that this procedure requires a long time, and is said to be less efficient [2].

Demand for orthopaedic medicine has increased over the past 20 years, based on the increasing number of patients each year [3]. Before surgery is performed, a conventional orthopaedic image identification is done by manually matching the articial implant image with a patient X-ray by an orthopaedic specialist. This method is a conventional method to determine the patient's implant size. But the manual or observational procedures require a long time to recognize the size of the patient's implant because the method used is repeated several times. Thus, the manual procedure should be changed to a digital and automated technique, or more accurately by using software. This technique helps the surgeon identify the appropriate patient's implant size digitally and automatically. Several studies have shown that a digital technique can improve the placement of a total hip implant [4-7].

## 2. SCOPE AND OBJECTIVES

The main objective of this research is to produce a technique which can automatically recognize the size of patient's acetabular implant in total hip replacement surgery. Particularly, in order to achieve these goals, the main objectives of the research that has been identified are:

   i. Studying and designing techniques for the recognition of hip replacement acetabular implant size.
   ii. Implementing the techniques developed.

## 3. RESEARCH BACKGROUND

An acetabular implant size recognition technique to be developed involves a number of things that need to be studied carefully. The technique could be developed to recognize the size of the acetabular implant accurately and effectively. In this section, the techniques and other related matters will be discussed.

### 3.1 Computer Aided Design

Computer Aided Design, better known as CAD, is defined as a system that uses computers to assist in designing and sketching work [8]. CAD is very helpful in some areas such as the use of appropriate scale, object manipulation, display and printing. Currently, there are various types of CAD software such as AutoCAD, Solidworks, CATIA, and MasterCAM. In this study, CAD will be used to design the related implant. Examples of implant that were designed using AutoCAD software can be seen in Figure 1.

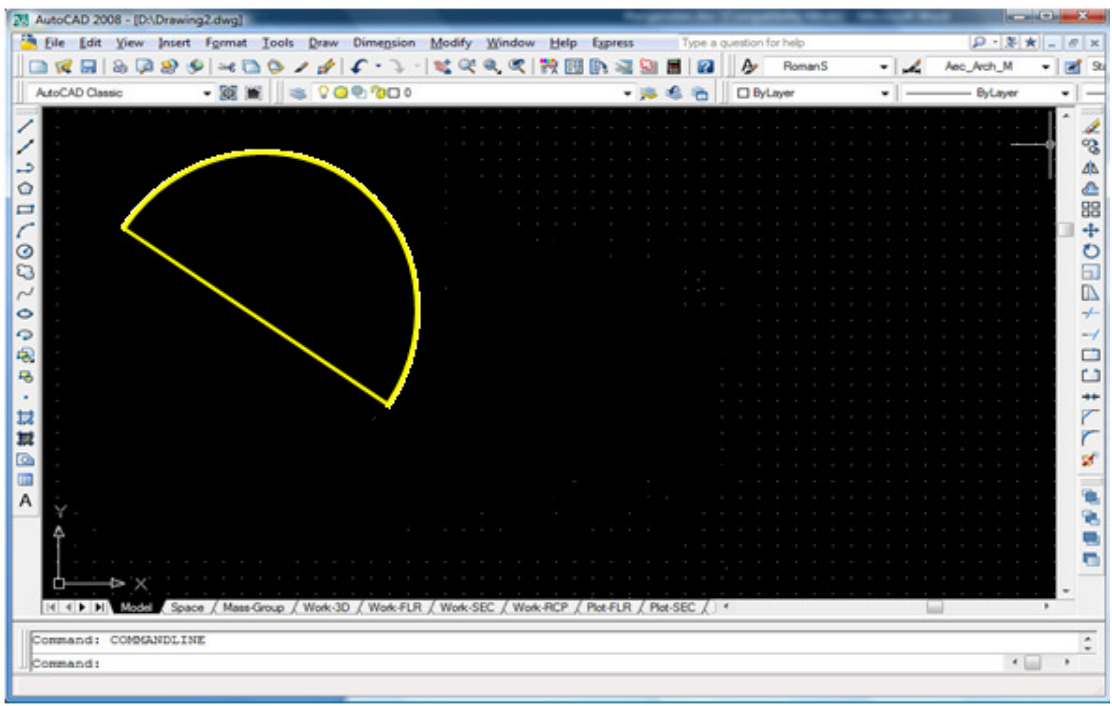

Figure 1. Acetabular implant design

### 3.2 Digital X-Ray Images

X-ray is a type of radiation used in medical images for diagnosing diseases like cancer and fractures [9]. The radiologist takes X-ray images by putting on an X-ray to an opposite source of the part that needs to be imaged or convert it into films. Then, the image can be generated into films or stored digitally. The difference between a digital X-ray and a general X-ray is that the output given in the first case is in a digital form while the general X-ray is in films. A digital X-ray may be edited and stored in a computer database. While a general x-ray only provides a negative film output as a reference. An example of a digital X-ray can be seen in Figure 2. An x-ray image is stored in JPEG format using software MedWeb [10].

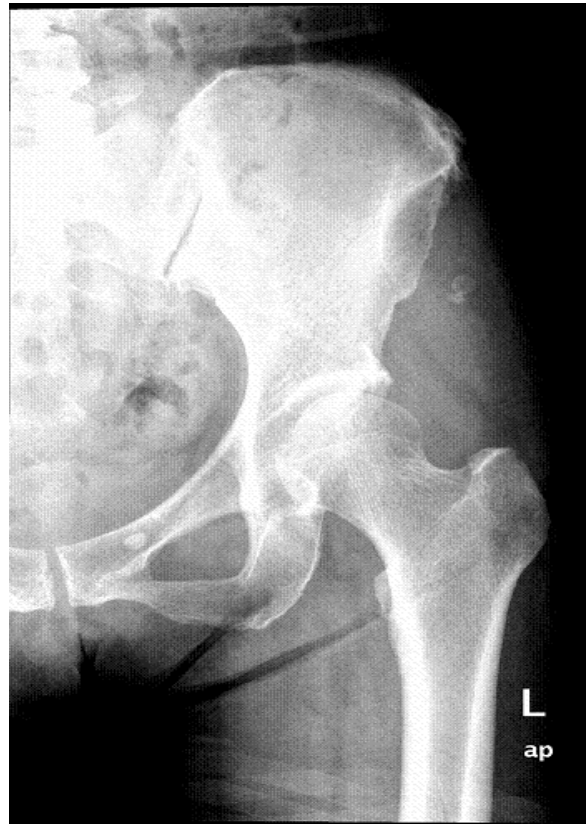

Figure 2. Digital X-ray

Each digital x-ray image has a different resolution depending on the amount of image compression. The purpose compression is done is to reduce the file size so that the use of memory space can be reduced, and to accelerate the transfer of files. The disadvantage of compression is that it can affect the image quality. Therefore, an appropriate resolution and techniques in scaling digital X-Ray images should be produced because the size of the implant identification technique to be developed depends on the accuracy of the image size of the patient's bone.

### 3.3 Image and Object Recognition Techniques

The image recognition operation is useful in helping to edit operation and object manipulation interactively [11]. The object is polygon-based, and the points in the polygon can form a polygon area. The polygon area can be identified by testing the position of the mouse on the digital X-ray images area. Object recognition operations can be done by using the dots on the X-ray images. This technique is very important because it can determine the optimum acetabular implant size.

### 3.4 Total Hip Replacement (THR)

Every joint disease, whether inflammatory or trauma, - if the disease process is allowed to continue - will cause erosion of cartilage and joint damage. When this happens, a patient may complain of pain, swelling, deformity and instability of joints involved. There are several methods of treatment for damaged joints. The first step is a move in the conservative treatment. If this fails, then the next method of treatment is surgery. The most effective surgical method is the total hip replacement THR surgery. This surgery involves an artificial joint that replaces the original which is damaged. THR is a surgical procedure in which the hip joint is replaced by an implant. THR surgery can be performed as a total replacement or a hemi (half) replacement [12]. THR consists of replacing both the femoral head and the acetabulum.

#### 3.4.1 Acetabulum

The acetabulum (see figure 3) is a concave surface of the pelvis. The head of the femur meets with the pelvis at the acetabulum, forming the hip joint [12].

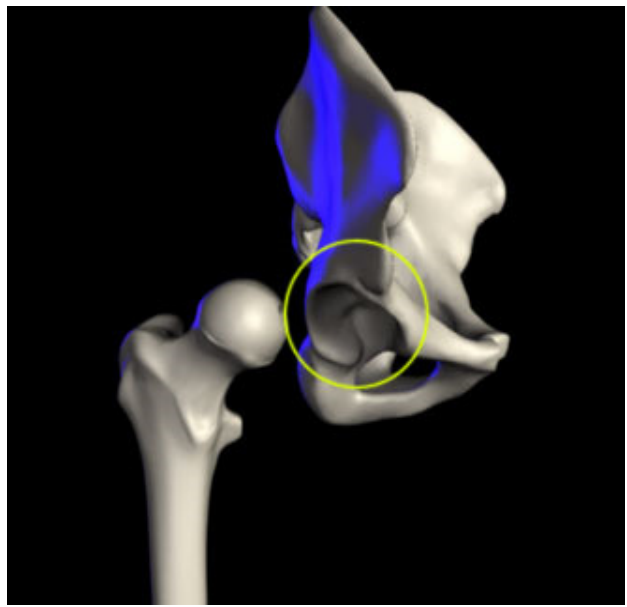

Figure 3. The acetabulum

#### 3.4.2 Acetabular

The acetabular implant is the component which is placed into the acetabulum (hip socket). Bone and cartilage are removed from the acetabulum, and the acetabular implant is attached using cement or friction. Some acetabular cups are one piece, others are modular. One piece shells are either polyethylene or metal, they have their articular surface machined on the inside surface of the cup and do not rely on a locking mechanism to hold a liner inplace [12]. Figure 4 shows the acetabular implant used in THR, while figure 5 shows the acetabular position after THR is performed [13].

239

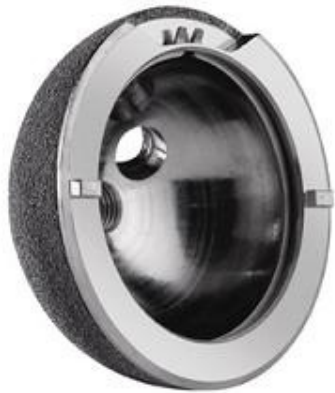

Figure 4. Acetabular implant

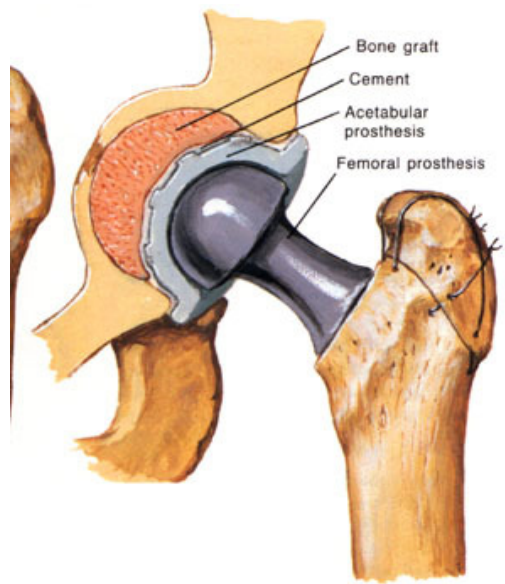

Figure 5. Acetabular position after THR is performed

### 3.4.3 Acetabular Conventional Templating

Prior to surgery, the template preparation process is very important because it is a screening process in which an orthopaedic specialist can determine the optimal implant size before surgery. The orthopaedic specialist will do the adjustment process of the implant templates first before bringing the results into the operating room. In the operating room, the orthopaedic specialist will make the final selection of the appropriate implant size. The use of implants in the template customization process on the X-ray image of a patient before surgery can provide important information to the orthopaedic specialist in determining the size of the implant to be used [14]. By using the observational or manual method, any error in template transformation such as rotating and scaling while recording the patient's hip bone radiographs prior to the surgery will lead to errors in determining the size of implant. Thus, the templating process

240

International Journal of Computer Science & Information Technology (IJCSIT), Vol 3, No 2, April 2011

should be done cautiously. Fig. 6 shows the process of manual implant templating for acetabular component [15].

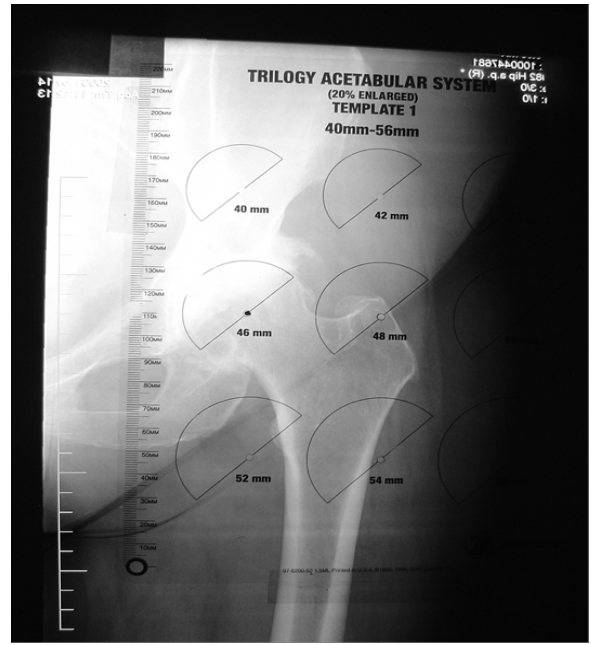

Figure 6. Manual acetabular implant templating

## 4. MATERIALS AND METHOD

The main activity to implement is to analyze the existing techniques for the implant recognition process. The technique analysis aims to identify similar techniques or that have been implemented or are being implemented by other researchers around the world. The process of producing a new technique for the implant size recognition is to take into account the improved methodology to be adopted on the latest techniques. the new method that has the potential to be developed is the automated recognition of the digital acetabular implant size. This means that the surgeons do not have to use a *trial and error* method to select the appropriate implant size [16].

The second objective of the research is to implement the new techniques produced by developing a system that consists of modules of the new technique. This aims to ensure that new techniques will be developed to be implemented effectively. In addition, the system will be developed to include other modules such as the transformation of the image, measuring the size and angle, and so forth. An appropriate programming language is to be used for this purpose. After a new technique has been implemented, the next phase is to do testing and comparison. The aim is to ensure that this new technique can effectively recognize the implant size, and to assist physicians in carrying out the pre-operative process before the surgery in order for it to be more efficient and organized.

### 4.1 Acetabular Implant Template

Fig. 7 shows the acetabular implant template used by PPUKM surgeons in the THR preoperative planning procedure.

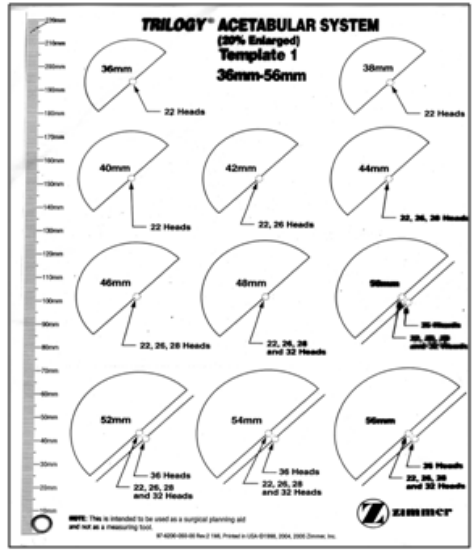

Figure 7. Acetabular template

## 4.2 Digital Implant

Digital implants produced using Autocad 2008 and Photoshop software can be seen in figure 8(a) and figure 8(b).

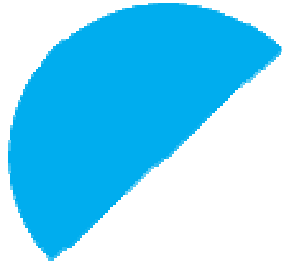

Figure 8(a). Digital acetabular implant (left hip)

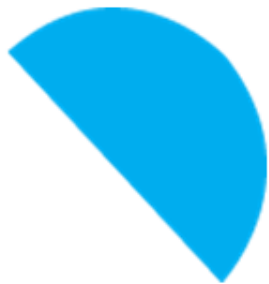

Figure 8(b). Digital acetabular implant (right hip)

## 4.3 Acetabular Implant Detection Technique

The process of designing and producing new techniques of implant detection is taking into consideration the method of improvement which will be implemented on the latest techniques.

242

Through literature studies, most of the implant detection techniques used a *trial and error* method [2,3,5,6]. This means that surgeons need to try several sizes to get the optimum implant. One of the potential new methods to be produced is the automated detection method.

Detection techniques that will be generated are based on observational and computerized techniques in which the implant template will be matched to the X-ray images. A *Trial and error* method will be repeated several times until the appropriate size is found. The proposed new technique has the ability to detect the size of the acetabular implant automatically. By using the concept of distance between two points, acetabular implants will be selected based on the patient's acetabulum diameter size. Fig. 9 shows the proposed algorithm for the acetabular implant size recognition technique.

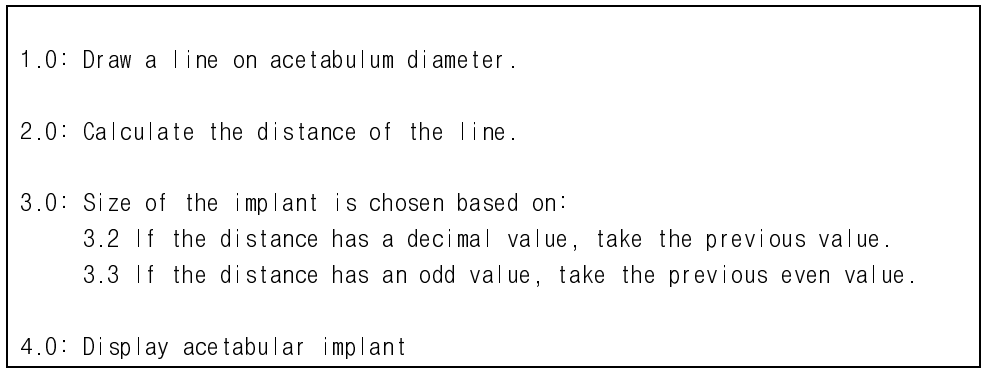

```
1.0: Draw a line on acetabulum diameter.

2.0: Calculate the distance of the line.

3.0: Size of the implant is chosen based on:
     3.2 If the distance has a decimal value, take the previous value.
     3.3 If the distance has an odd value, take the previous even value.

4.0: Display acetabular implant
```

Figure 9. Acetabular implant size recognition algorithm

In the algorithm shown in figure 9, the user must draw a line on the acetabulum diameter. Line distance will be calculated, and the size of the implant will be determined by the acetabulum diameter size. Sizing method involves three conditions, namely, if the size of the implant is smaller than 36 mm or larger than 80 mm, then the implant will not be displayed. This is because the TRILOGY acetabular system implant comes only in size from 36 mm to 80 mm. In addition, if the distance has a decimal value, the previous value will be taken. For example, in figure 10, the distance of the line drawn on the acetabulum diameter is 64.15 mm, therefore the value will be taken as 64 mm.

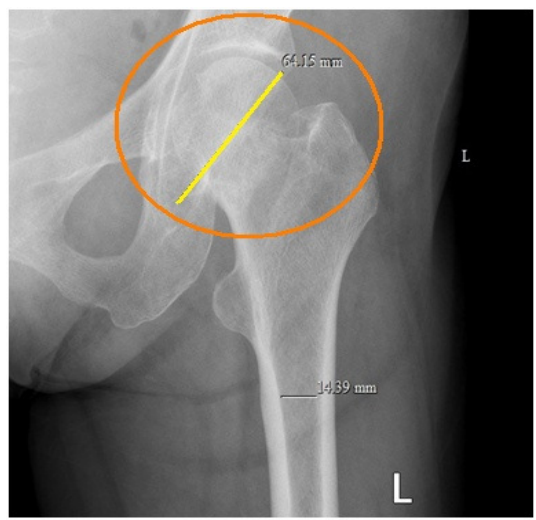

Figure 10. Line drawn on acetabulum

International Journal of Computer Science & Information Technology (IJCSIT), Vol 3, No 2, April 2011

## 5. RESULTS AND DISCUSSION

Based on the algorithm in Fig. 9, the first thing the user needs to do is draw a straight line on the patient's acetabulum diameter. Fig. 11 shows how to draw the line.

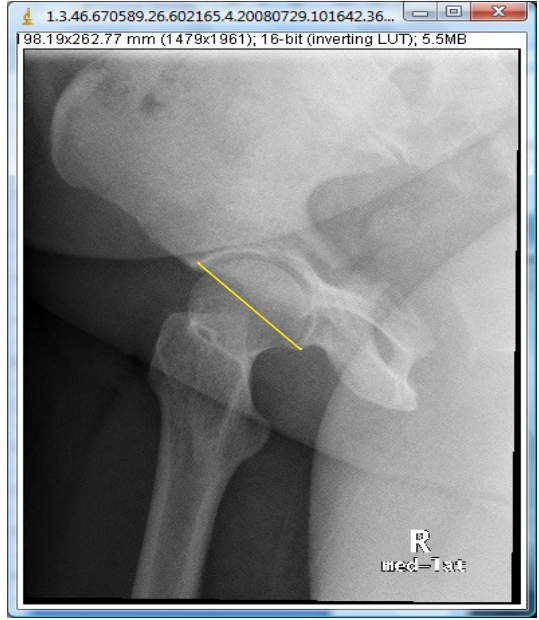

Figure 11. Straight line drawn on the acetabulum diameter

The distance between two points on the line will be calculated to determine the size of the acetabular implant. In Fig. 11, the distance drawn on the acetabulum diameter was 58.28 mm, so the value to be taken is 58 mm. An acetabular implant with 58 mm size will be displayed automatically on the X-ray image (see fig. 12).

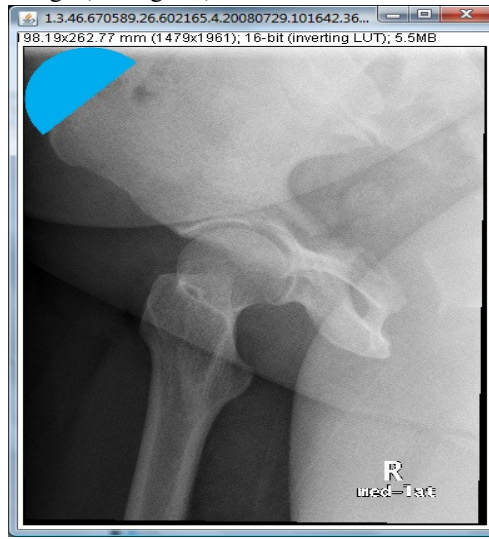

Figure 12. Acetabular implant default position

The size of the acetabular supplied by the PPUKM is an even number, so if there is an odd number, the even value before the odd value will be taken as the size of the acetabular implant. For example, if the distance drawn was 59.25 mm, the value to be taken is 58 mm. Table 1 shows the examples how to determine the acetabular size based on the distance between two points.

Table 1. Acetabular size based on distance between two points

| No. | Distance (mm) | Acetabular size (mm) |
| --- | --- | --- |
| 1 | 48.58 | 48 |
| 2 | 57.45 | 56 |
| 3 | 58.15 | 58 |
| 4 | 53.36 | 52 |
| 5 | 66.45 | 66 |
| 6 | 69.13 | 68 |
| 7 | 72.78 | 72 |
| 8 | 77.67 | 76 |

In Fig. 12, it can be seen that the position of the acetabular implant is still not in the optimum position. The implant should be translated and rotated to get the optimum position. Figure 13(a) shows the implant which is translated and rotated using the geometric transformation algorithm [17], while figure 13(b) shows the information of implant and patient.

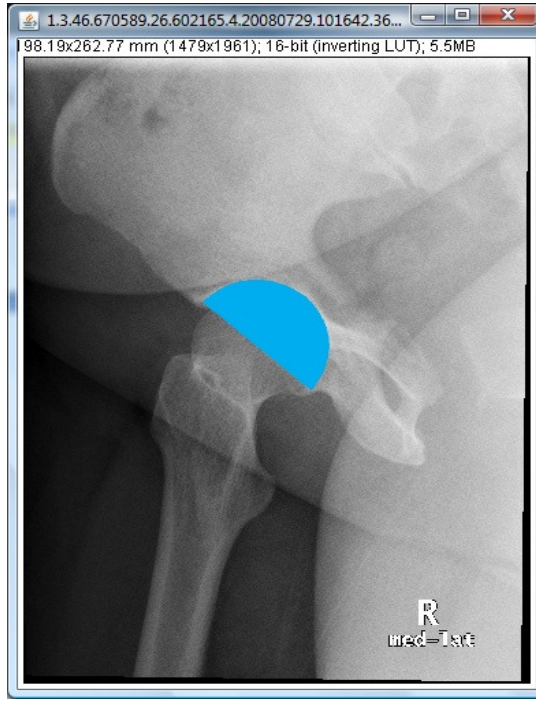

Figure 13(a). Implant that has been translated and rotated

245

Figure 13(b). Acetabular and patient's information

To test the accuracy and utility of the acetabular implant size recognition technique, an experiment was conducted with assistance from a surgeon for the observational (manual) method to determine the acetabular implant size. The results by the observational approach were compared to the results produced by our automated technique. The testing recorded the acetabular implant size to be used and the time taken by both methods. Ten randomly selected X-rays of unidentified patients were used for templating for both techniques. The difference between the two sizes were calculated and shown in table 2.

Table 2. Observational vs Digital

| Patient | Stem Size (Observational) | Stem Size (Digital) | Size Difference |
|---|---|---|---|
| 1 | 48 | 48 | 0 |
| 2 | 54 | 52 | ±2 |
| 3 | 50 | 50 | 0 |
| 4 | 52 | 52 | 0 |
| 5 | 46 | 46 | 0 |
| 6 | 48 | 46 | ±2 |
| 7 | 52 | 52 | 0 |
| 8 | 58 | 54 | ±4 |
| 9 | 56 | 54 | ±2 |
| 10 | 54 | 54 | 0 |

It is evident that the new technique yields very close results to those obtained by the observational method in nine studies (90%). The difference, if any, is also within the error of clinically acceptable range (± 2 size determined by the surgeon) obtained by the observational templating method (only one study (10%) is outside of clinically acceptable range). In addition, the study also demonstrated that the average time taken for acetabular implant templating in total hip replacement preoperative planning using an automated technique was much less than using the observational method.

## 6. CONCLUSIONS

Preoperative templating has been useful to determine the optimum implant size in total hip replacement (THR). With classical tracing paper now obsolete, we have developed a new technique to undertake the templating procedure with a digital acetabular implant and an X-ray. The digital implant provides several advantages for THR surgery. Compared to the observational method in which the surgeon uses a template manually and places it on the patient's X-ray, the use of the digital implant not only saves time, but also can reduce the error due to consistency difference when making adjustments to a patient's implant size [14].

This new technique offers a simple solution to the problem of using the observational method in THR. The technique allows users to choose the acetabular implant automatically on computer prior to surgery, based on the diameter of the patient's acetabulum size. The new proposed acetabular implant detection technique also provides user-friendly and accurate computer programming surgical planning. In addition, the average time taken for the acetabular implant templating process in THR using an automated technique was much less than using the observational method.

## ACKNOWLEDGEMENT

This research project was conducted in collaboration with Dr. Abd Yazid Mohd Kassim, Dr. Hamzaini Abd Hamid and Dr Nor Hazla Haflah from the Department of Orthopaedics and Traumalogy, Medical Centre of Universiti Kebangsaan Malaysia. This department has provided medical image data (DICOM) to be used in this research. This research is also funded by University Grants UKM-OUP-ICT-35-179/2009 and UKM-GUP-TMK-07-01-035.

**Authors**

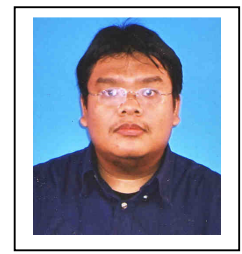

Azrulhizam Shapi'i is currently pursuing Ph.D in Industrial Computing at School of Information Technology, Universiti Kebangsaan Malaysia. He is working as Lecturer in the School of Information Technology, Faculty of Information Science and Technology, University Kebangsaan Malaysia. His research areas of interest include Computer Aided Design, Medical Imaging, Computer Aided Medical System and Programming.

Dr. Riza Sulaiman did his Msc from University of Portmouth, UK and Ph.D from University of Canterbury, New Zealand. His specializations include Computer Aided Design (CAD), Medical Imaging and Robots Simulation. He is working as Associate Professor in the School of Information Technology , Faculty of Information Science and Technology, University Kebangsaan Malaysia.

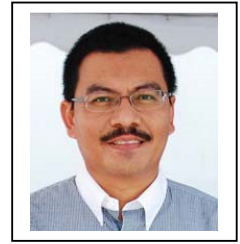

Dr. Khatim Hasan did his Msc from Universiti Kebangsaan Malaysia, and Ph.D from University Putra Malaysia. His specializations include Scientific Computing and Statistical Analysis. He is working as Associate Professor in the School of Information Technology , Faculty of Information Science and Technology, University Kebangsaan Malaysia. He has published 108 papers in reviewed journals and proceedings, 3 books and chapter in books, 12 in non-reviewed proceedings and 24 technical reports.

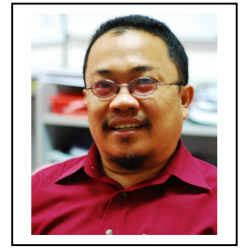

Dr. Abdul Yazid Mohd Kassim is an orthopaedic surgeon from Department of Orthopaedic and Traumalogy, Medical Centre of Universiti Kebangsaan Malaysia (PPUKM).

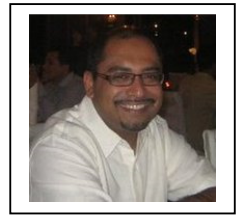